\documentclass[letterpaper, 10 pt, conference]{ieeeconf}  % Comment this line out if you need a4paper

\IEEEoverridecommandlockouts                              % This command is only needed if 
\usepackage{graphicx}
\usepackage{booktabs}
\usepackage {tablefootnote}
\usepackage{amssymb} % for \checkmark

\usepackage{amsmath}
\usepackage{algorithmic}
\usepackage{algorithm}
\usepackage{siunitx}
\usepackage{multirow}
\usepackage{url}
\usepackage[hang,small,bf]{caption}
\usepackage[subrefformat=parens]{subcaption}
\title{\LARGE \bf
MRHaD: Mixed Reality-based Hand-Drawn Map Editing Interface \\ for Mobile Robot Navigation}

\author{Takumi Taki$^{1\dag}$, Masato Kobayashi$^{1,2,3\dag*}$, Eduardo Iglesius$^{1}$, Naoya Chiba$^{1,2}$, Shizuka Shirai$^{1,2}$, Yuki Uranishi$^{1,2}$ % <-this % stops a space
\thanks{
${\dag}$ Equal Contribution,
$^{1}$ Graduate School of Information Science and Technology, The University of Osaka, $^{2}$ D3 Center, The University of Osaka, $^{3}$ Graduate School of Maritime Sciences, Kobe University, * corresponding author: kobayashi.masato.cmc@osaka-u.ac.jp}
}

\begin{document}

\maketitle
\thispagestyle{empty}
\pagestyle{empty}

%%%%%%%%%%%%%%%%%%%%%%%%%%%%%%%%%%%%%%%%%%%%%%%%%%%%%%%%%%%%%%%%%%%%%%%%%%%%%%%%
\begin{abstract}
Mobile robot navigation systems are increasingly relied upon in dynamic and complex environments, yet they often struggle with map inaccuracies and the resulting inefficient path planning. This paper presents MRHaD, a Mixed Reality-based Hand-drawn Map Editing Interface that enables intuitive, real-time map modifications through natural hand gestures. By integrating the MR head-mounted display with the robotic navigation system, operators can directly create hand-drawn restricted zones (HRZ), thereby bridging the gap between 2D map representations and the real-world environment. Comparative experiments against conventional 2D editing methods demonstrate that MRHaD significantly improves editing efficiency, map accuracy, and overall usability, contributing to safer and more efficient mobile robot operations. The proposed approach provides a robust technical foundation for advancing human-robot collaboration and establishing innovative interaction models that enhance the hybrid future of robotics and human society.
For additional material, please check: \url{https://mertcookimg.github.io/mrhad/}
\end{abstract}

%%%%%%%%%%%%%%%%%%%%%%%%%%%%%%%%%%%%%%%%%%%%%%%%%%%%%%%%%%%%%%%%%%%%%%%%%%%%%%%%

\section{INTRODUCTION}
Recent advances in autonomous mobile robots have opened up new opportunities for human-robot collaboration in various application domains, including logistics, healthcare, and public spaces \cite{zghair2021one, eriksen2023understanding, herath2023robots}.
Typically, these robots use pre-constructed environmental maps and dynamically adjust their paths based on real-time environmental sensing with various onboard sensors. Path planning methods are generally divided into two categories: global planning and local planning \cite{kobayashi2022dwv}. Global planning computes an optimal path from the start to the goal based on an environmental map prior to movement, whereas local planning utilizes real-time sensor data (e.g., LiDAR or camera feeds) to detect obstacles during operation and initiate evasive actions as needed.

However, discrepancies between pre-constructed maps and real-world settings, such as new or undetected obstacles, can cause inefficient detours or safety risks. 
For instance, global planning may create unnecessary detours to avoid obstacles that no longer exist or fail to accommodate newly introduced obstacles, thereby raising the risk of collisions. Similarly, in local planning, obstacles that are difficult to detect, such as those with low reflectivity or minimal height, may elude timely detection, limiting the robot’s ability to navigate safely.
These challenges become particularly critical when robots share dynamic and unpredictable environments with humans, highlighting the need for a robust, real-time, and human-centered approach to map modification.

Numerous approaches for map updates have been explored, including autonomous robot map reconstruction by the robot \cite{filipenko2018comparison} and manual intervention by human operators \cite{sidaoui2018human}. However, autonomous reconstruction is constrained by sensor accuracy, whereas manual editing can impose a high cognitive burden due to the lack of an intuitive relationship between the map and the real environment.

\begin{figure}[t]
    \centering
    \includegraphics[keepaspectratio, width=0.85\linewidth]{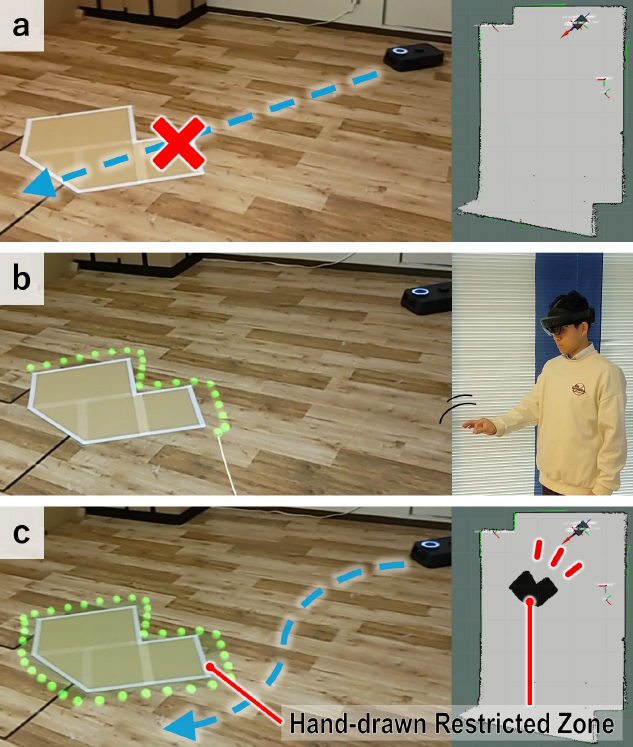}
    \caption{Overview of MRHaD system. (a) An obstacle not reflected on the map is not considered in path planning. (b) The operator is drawing Hand-drawn Restricted Zone (HRZ) through MR hand gesture. (c) The highlighted area is immediately incorporated into the navigation map as HRZ.}
    \label{fig:teser}
\end{figure}

This paper aims to address these limitations by introducing a map editing system that combines Mixed Reality (MR) technology with hand gesture operations.
An overview of our proposed system is shown in Fig.~\ref{fig:teser}.
The key feature of our system is the concept of ``Hand-drawn Restricted Zone (HRZ)'', a user-defined polygonal region that prevents the robot from entering specific areas. Using MR head-mounted display (MR-HMD), operators can visualize their surroundings and draw HRZ directly on the floor corresponding to obstacles in the real environment.
This lowers the cognitive load required for map editing, as the task of correlating 2D map data with 3D physical space becomes more intuitive.
Furthermore, the edited map with newly drawn or adjusted HRZ is immediately updated in the robot's navigation and map system, supporting efficient path planning without the detours or collisions that can arise from map inaccuracies. By optimizing the real-time creation, modification, and deletion of HRZ, our system aims to improve both the safety and efficiency of mobile robot navigation.

Our contributions are summarized as follows: 
\begin{itemize}
    \item We proposed MRHaD, a novel MR-based hand-drawn map editing approach that enables the intuitive creation and deletion of HRZ through hand gestures.
    \item MRHaD supported the reduction of cognitive load by integrating real-environment visualization and interactive editing on MR-HMD, eliminating the mental gap between 2D map data and the physical world.
    \item We conducted comprehensive experimental validation showing that our method substantially outperforms a traditional 2D-based system in efficiency, accuracy, and usability.
\end{itemize}

This paper consists of six sections. Section II explains  Related Works.
Sections III and IV propose MRHaD System Design and System Implementation.
Section V confirms the method's efficiency based on the experimental results.
Section VI concludes this paper.

\section{RELATED WORK}\label{sec:RW}
\subsection{Robot Navigation Interfaces}
Traditional navigation interfaces have primarily used 2D graphical user interfaces (GUIs) to display sensor data, such as camera, sonar, and LiDAR readings, and to allow users to specify goals or plan paths. Previous work introduced systems that overlay 2D maps with sensor-based environment information, improving situational awareness for human operators \cite{kawamura2003agent, olivares2003interface}. Subsequent studies explored more intuitive interactions, such as varying methods for setting destinations \cite{labonte2006pilot} and block-based approaches for control and path specification \cite{tiddi2018user}. Furthermore, RViz remains a popular tool, providing both 2D and 3D visualization of the robot’s environment \cite{kam2015rviz}.

Recently, MR-HMD has garnered interest for its ability to overlay virtual information onto the real environment. Some research has used MR-HMD to visualize robots through walls or occlusions \cite{gu2021seeing}, or to enable interaction through hand or pointer gestures \cite{quesada2022holo, gu2022ar, iglesius2024mrnab}. These MR-based interfaces primarily address navigation goal setting and real-time feedback rather than map editing.

Our paper focuses more on efficient map modification by defining HRZ, rather than merely commanding motion targets or visualizing the robot’s state. We utilize MR-HMD not only for guidance but also to enable direct, intuitive editing of spatial information used for navigation.
\begin{table}[tb]
    \centering
    \caption{Comparison with Other Map Editing Interfaces}
    \label{tab:related_works}
    \resizebox{0.48\textwidth}{!}{
    \begin{tabular}{lccccc}
        \toprule
        & \multicolumn{2}{c}{2D} & \multicolumn{3}{c}{MR} \\ \cmidrule(lr){2-3} \cmidrule(lr){4-6} 
        Index & \: I1 & I2 \: & \: I3 & I4 & MRHaD \: \\ \midrule
        Map Editing Accuracy & \: $\checkmark$ & $\checkmark$ & \: $\times$ & $\checkmark$ & $\checkmark$ \: \\
        Required Editing Time & \: $\times$ & $\times$ & \: $\checkmark$ & $\times$ & $\checkmark$ \: \\
        Real Environment Spatial Understanding & \: $\times$ & $\times$ & \: $\times$ & $\checkmark$ & $\checkmark$ \: \\
        \bottomrule
        \multicolumn{6}{l}{
        \begin{tabular}{l}
            \footnotesize I1: Sidaoui et al., 2018\cite{sidaoui2018human}, I2: Koide et al.\cite{koide2020interactive}, I3: Wu et al. \cite{wu2020mixed}, \\I4: Sidaoui et al., 2019\cite{sidaoui2019collaborative}, Puljiz et al.\cite{puljiz2020hololens}
        \end{tabular}
        }
    \end{tabular}
    }
\end{table}
\begin{figure}[t]
    \centering
    \includegraphics[keepaspectratio, width=0.8\linewidth]{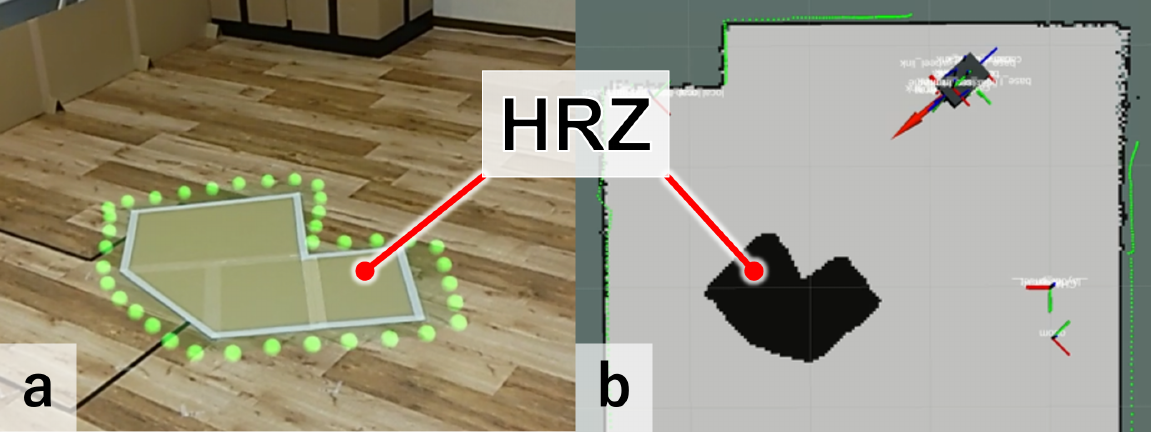}
    \caption{An example of HRZ. (a) shows HRZ displayed on HoloLens~2. (b) shows its reflection on ROS~2 cost map.}
    \label{fig:rzp_sample_ros2}
\end{figure}
\subsection{2D Map Editing Interfaces}
Alongside navigation interfaces, there has been significant effort in tools for editing robot-generated maps, usually derived from Simultaneous Localization and Mapping (SLAM). For example, Sidaoui et al.\ proposed iterative refinements of 2D occupancy grid maps, where human operators can remove noise or add obstacles that the robot’s sensors missed \cite{sidaoui2018human}. Similarly, Koide et al.\ introduced an interactive approach to correct map distortions and misalignment in 3D reconstructions through a conventional GUI \cite{koide2020interactive}.

While 2D map editing tools can significantly improve map accuracy and provide flexible oversight for operators, they typically require users to mentally transform between 2D view and the physical 3D setting. Our work tackles this inherent limitation by merging the editing process with an on-site MR visualization. Through this method, operators can inspect the real environment and immediately validate the accuracy of any edits they make on the map, thereby reducing cognitive load and potential user error.
\begin{figure*}[t]
    \centering
    \includegraphics[keepaspectratio, width=1\linewidth]{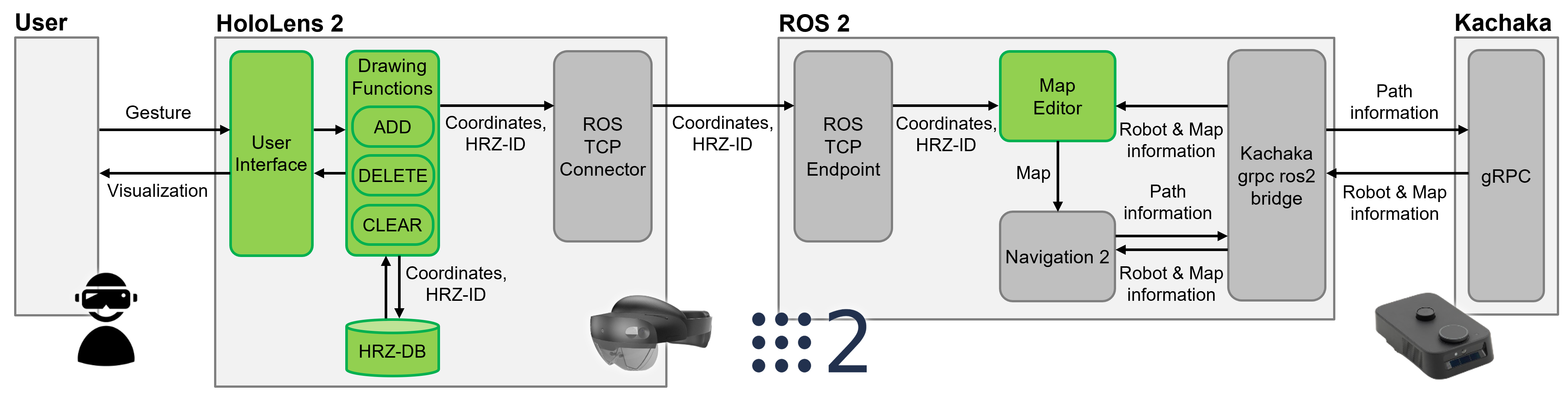}
    \caption{System Structure and Data Flow. Once the drawing operation is completed, the vertex coordinates and identification number (HRZ-ID) of the HRZ are sent to ROS~2. ROS~2 converts the received coordinates appropriately, stores them along with HRZ-ID, and then adds the HRZ to the map received from Kachaka. When HRZ is deleted, the corresponding HRZ-ID is sent from HoloLens~2 to ROS~2, which then removes only the specified HRZ from the map.}
    \label{fig:system_structure}
\end{figure*}
\subsection{MR-Based Map Editing Interfaces}
To make editing more intuitive, some researchers have turned to MR for direct manipulation of the environment map. Sidaoui et al.\ demonstrated collaborative, real-time correction of SLAM-generated maps through MR-HMD interface \cite{sidaoui2019collaborative, sidaoui2019slam}. Wu et al.\ introduced the placement of virtual obstacles on a 2D map displayed in MR-HMD \cite{wu2020mixed}, while Puljiz et al.\ used MR-HMD to overlay voxel maps onto the real environment, enabling users to mark safe zones or no-go areas \cite{puljiz2020hololens}.

However, the majority of these systems rely on detailed grid-level or voxel-level editing, which, although precise, can be time-consuming and cumbersome for large-scale adjustments. Moreover, broad modifications, such as drawing a new restricted zone, often require multiple steps or advanced spatial alignment.

Our approach seeks to streamline these processes by focusing on intuitive, freehand editing of HRZ.
Rather than manipulating low-level map cells, operators can outline “HRZ” in a single hand gesture, visually confirm placement via MR overlay, and immediately propagate changes to the robot’s global navigation map. This design enables faster and more flexible updates while reducing the cognitive burden of correlating 2D map data with 3D real-world surroundings.
To clarify the advantages of our approach, Table~\ref{tab:related_works} summarizes the comparison between our interface and other relevant interfaces.

\section{SYSTEM DESIGN}\label{sec:SD}

\begin{figure}[t]
    \centering
    \includegraphics[keepaspectratio, width=1\linewidth]{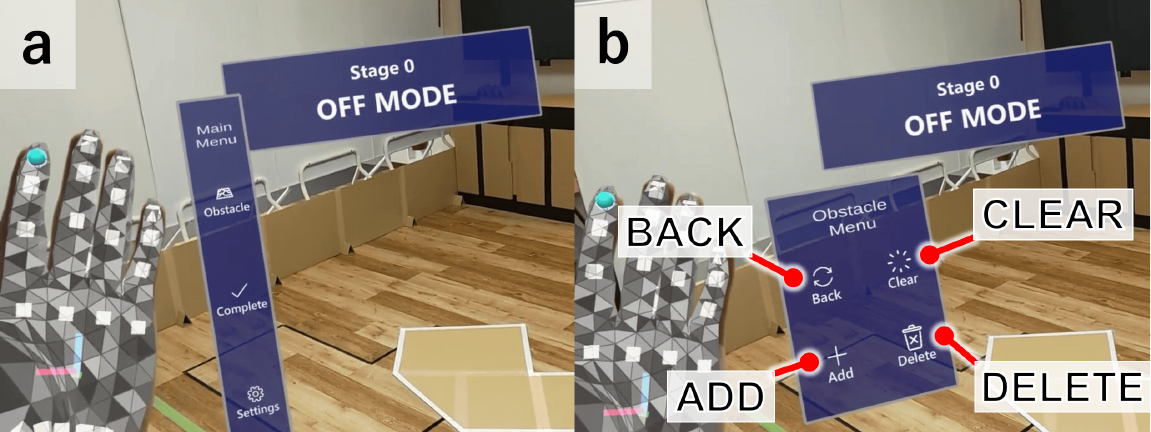}
    \caption{Menu  Panel. (a) Initially, the main menu is displayed. (b) Pressing the Operation Button switches to the drawing operation menu.}
    \label{fig:mode_main_obstacle}
\end{figure}

The user wears HoloLens~2 and draws Hand-drawn Restricted Zones (HRZ) using hand gestures. The drawn HRZ is visualized as a green polygon on HoloLens~2. Fig.~\ref{fig:rzp_sample_ros2} shows an example of HRZ displayed on HoloLens~2 and its reflection on the cost map within the Robot Operating System~2 (ROS~2).

In ROS~2, the addition of HRZ is realized by connecting the received vertex coordinates to form a polygon and marking its edges and interior cells as occupied. Deletion is achieved by resetting the map and reapplying all HRZ except the one marked for deletion.  HRZ-ID and vertex coordinates of each HRZ are stored in an internal database of HoloLens~2, allowing the system to restore them upon restart.

The structure and data flow of the proposed system are shown in Fig.~\ref{fig:system_structure}. The system consists of a user, HoloLens~2, ROS~2, and the mobile robot, Kachaka. The newly developed components in this paper are shown in green.

The proposed system's drawing operations consist of three functions: ``ADD'', ``DELETE'', and ``CLEAR''. The HoloLens~2 drawing system includes these drawing modes as well as an ``OFF mode'' where no drawing function is active.
All of the functions can be accessed from the menu panel shown in Fig.~\ref{fig:mode_main_obstacle}.

\subsection{ADD Function}\label{sec:SD_function_add}
\begin{figure}[t]
    \begin{minipage}[b]{1\linewidth}
        \centering
        \includegraphics[keepaspectratio, width=1\linewidth]{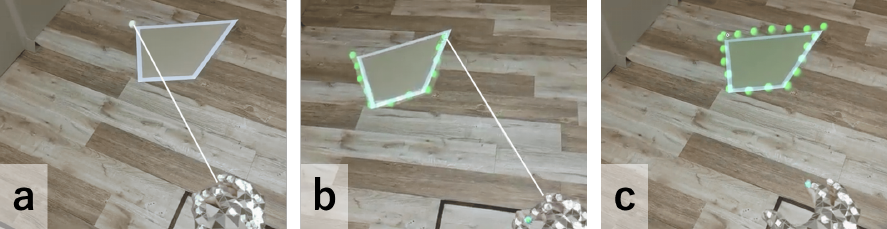}
        \vspace{-5mm}
        \caption{ADD Operation Procedure. (a) The user extends a hand with the palm facing downward, displaying a white cursor on the floor. A pinch gesture places a sphere at the cursor position. (b) Moving the hand draws a polyline outlining the desired HRZ. (c) Releasing the pinch connects the start and end points to form a closed polygon, which is then filled with green to indicate the HRZ.}
        \label{fig:operation_add}
    \end{minipage}
\end{figure}

The ADD function allows the user to create new HRZ.
When the ADD button is selected from the drawing operation menu, the system switches to ADD mode.
In this mode, the user can draw a polygon, which serves as the HRZ, using cursor operations.
Once the HRZ object is generated, its HRZ-ID and coordinate information are added to the database and transmitted to ROS~2.
The ADD operation procedure is illustrated in Fig.~\ref{fig:operation_add}.

\subsection{DELETE Function}\label{sec:SD_function_delete}
\begin{figure}[t]
    \begin{minipage}[b]{1\linewidth}
        \centering
        \includegraphics[keepaspectratio, width=1\linewidth]{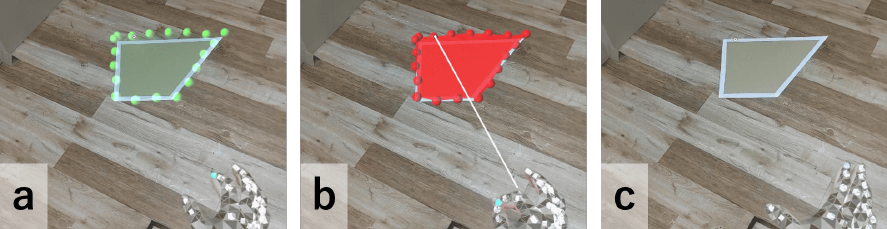}
        \vspace{-5mm}
        \caption{DELETE Operation Procedure. (a) The user aligns the cursor with one of the spherical vertices of the HRZ to be deleted. (b) When the user pinches the thumb and index finger together, the entire HRZ object turns red, indicating selection. (c) The user then releases the pinch, and the HRZ is deleted.}
        \label{fig:operation_delete}
    \end{minipage}
\end{figure}

The DELETE function allows the user to select and remove a specific HRZ.
When the DELETE button is selected from the drawing operation menu, the system switches to DELETE mode.
In this mode, the user can delete HRZ by clicking on it with a pinch gesture (pinching and releasing the thumb and index finger).
After deletion, the corresponding HRZ object is removed from the database, and its HRZ-ID is sent to ROS~2.
The DELETE operation procedure is illustrated in Fig.~\ref{fig:operation_delete}.

\subsection{CLEAR Function}\label{sec:SD_function_clear}
\begin{figure}[t]
    \begin{minipage}[b]{1\linewidth}
        \centering
        \includegraphics[keepaspectratio, width=1\linewidth]{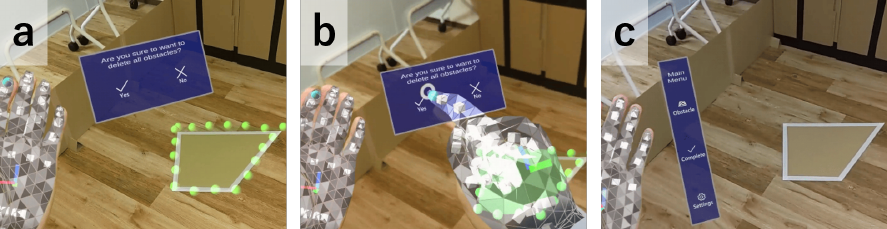}
        \vspace{-4mm}
        \caption{CLEAR Operation Procedure. (a) Upon entering CLEAR mode, a confirmation pop-up appears, asking whether to delete all HRZ. (b) Pressing the ``Yes'' button deletes all HRZ and returns the system to OFF mode. (c) Pressing the ``No'' button cancels the operation, keeping the HRZ unchanged and returning the system to OFF mode.}
        \label{fig:operation_clear}
        % \vspace{3mm}
    \end{minipage}
\end{figure}

The CLEAR function removes all existing HRZ.
When the CLEAR button is selected from the drawing operation menu, the system switches to CLEAR mode, displaying a final confirmation popup.
Pressing the ``Yes'' button deletes all HRZ and returns the system to OFF mode.
Simultaneously, all HRZ objects are removed from the database, and an initialization command is sent to ROS~2.
The CLEAR operation procedure is illustrated in Fig.~\ref{fig:operation_clear}.

\section{SYSTEM IMPLEMENTATION}\label{sec:SI}
\subsection{Communication Method}
The proposed system enables communication between Unity and ROS~2, and between ROS~2 and Kachaka.
For communication between Unity and ROS~2, ROS-TCP-Connector, and ROS-TCP-Endpoint are utilized. On the ROS~2 side, ROS-TCP-Endpoint functions as a TCP server, waiting for connections from Unity. On the Unity side, ROS-TCP-Connector is imported, and a ROS connection object is added to the scene. By setting the IP address and port number of the computer running ROS~2, message transmission and reception via topics and services become possible.

For communication between ROS~2 and Kachaka, kachaka-api is used. This API, based on gRPC, allows external control of Kachaka and retrieval of its information.

\subsection{Coordinate Alignment between Systems}
The system synchronizes the Unity and ROS~2 coordinate systems using HoloLens~2. To achieve this, the QR code recognition technology of Vuforia Engine is employed. By placing a QR code at the origin of the ROS~2 coordinate system in the real environment and a corresponding QR code object at the origin of the Unity coordinate system, HoloLens~2 can recognize the QR code and align the two coordinate systems.

\subsection{Coordinate Data Management}
The proposed system stores the ID and vertex coordinates of HRZ in a database, allowing the restoration of previously drawn content upon system restart. The database consists of a JSON file within the Unity project and a program that updates its content. The JSON file records each HRZ object's ID, coordinates (based on a reference QR code), rotation angle, and an array of vertex coordinates.
The pseudocode representing the algorithm for managing coordinate data is shown in Algorithm \ref{alg:database}.
At system startup, this JSON file is loaded, and the HRZ data is transmitted to ROS~2. When HRZ is added or deleted, the database is updated, and the latest state is saved by overwriting the JSON file.

\begin{algorithm}[tb]
    \small
    \caption{HRZ Data Management Process}
    \label{alg:database}
    \begin{algorithmic}[1]
    \STATE $database \leftarrow$ LoadJsonFile()
    \STATE InstantiateAllHRZ($database$)
    \WHILE{system is running}
    \IF{$currentMode$ is ADD}
    \STATE $database \leftarrow$ LoadJsonFile()
    % \STATE SaveDatabase($database$, $hrz\_id$, $coordinates$, $rotation$)
    \STATE SaveDatabase($database$, $hrz\_id$, $coordinates$, \\
    \hspace{2em}$rotation$)
    \STATE OverwriteJsonFile($database$)
    \ELSIF{$currentMode$ is DELETE}
    \STATE $database \leftarrow$ LoadJsonFile()
    \STATE DeleteDatabase($database$, $hrz\_id$)
    \STATE OverwriteJsonFile($database$)
    \ELSIF{$currentMode$ is CLEAR}
    \STATE $database \leftarrow$ GenerateEmptyJsonFile()
    \STATE OverwriteJsonFile($database$)
    \ENDIF
    \ENDWHILE
    \end{algorithmic}
\end{algorithm}

\subsection{Addition and Deletion of HRZ in the Map}
% This section describes the process of adding and deleting HRZ in ROS~2.
ROS~2 maintains a robot-generated cost map and an HRZ list containing HRZ-IDs and vertex coordinates. In the cost map, each cell is assigned one of three cost states: ``occupied'', ``free'', or ``unknown''.
The cost map is updated based on HRZ additions and deletions instructed by HoloLens~2.
Information for adding and deleting HRZ is transmitted from HoloLens~2 to ROS~2 through separate topics.

\subsubsection{ADD}
In the ADD operation, HoloLens~2 sends a message via a topic to ROS~2, containing HRZ-ID and a list of vertex coordinates for the HRZ to be added.
HRZ-ID is an integer value greater than or equal to 1.
The transmitted coordinates are expressed in the ROS~2 coordinate system.
ROS~2 subscribes to this topic and retrieves the received HRZ data.
ROS~2 converts the received coordinates into the cost map's coordinate system and stores them in the HRZ list along with its HRZ-ID.
It then connects the stored coordinates to draw a polygon on the cost map and fills its interior, thereby adding the HRZ.
On the cost map, ROS~2 changes the cost status of cells located along the edges and inside the polygon formed by connecting the stored coordinates to the ``occupied'' state.

\subsubsection{DELETE}
In the DELETE operation, HoloLens~2 sends a message via a topic to ROS~2, containing the HRZ-ID of the HRZ to be deleted.
ROS~2 subscribes to this topic, recognizes the received message as the target HRZ-ID, and removes the corresponding HRZ’s coordinate data from the HRZ list.
The cost map is reverted to the original robot-generated cost map, and all remaining HRZs are redrawn using the same method as the ADD operation.

\subsubsection{CLEAR}
In the CLEAR operation, HoloLens~2 sends a value of 0 to ROS~2 via the same topic as DELETE.
Since HRZ-IDs are integers greater than or equal to 1, if ROS~2 receives a value of 0, it recognizes this as a CLEAR operation and resets the cost map.

\section{EXPERIMENT}\label{sec:EXP}
\begin{figure}[t]
    \centering
    \includegraphics[keepaspectratio, width=\linewidth]{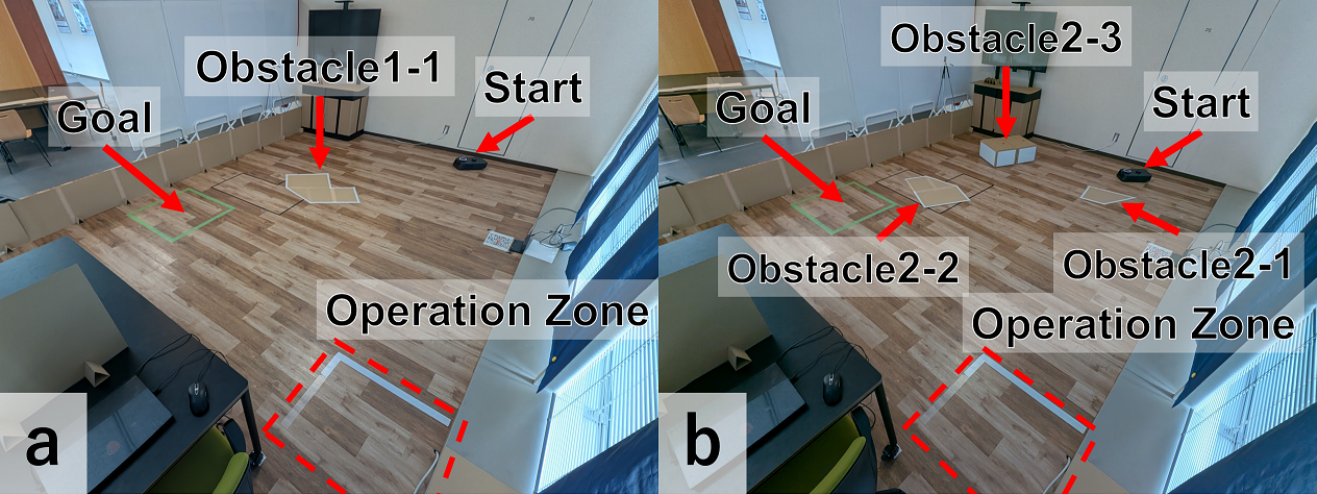}
    \caption{Experimental Environment. (a) Stage1: A single flat cardboard panel is used as the obstacle. (b) Stage2: Two flat cardboard panels and one box-shaped obstacle are arranged to form a more complex configuration.}
    \label{fig:experiment_environment}
\end{figure}
\begin{figure}[t]
    \centering
    \includegraphics[keepaspectratio, width=0.9\linewidth]{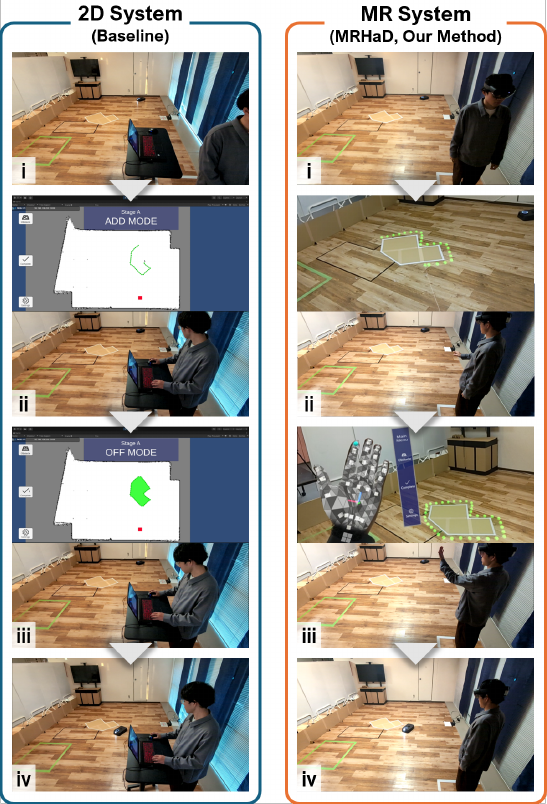}
    \caption{Overview of the Experimental Task Flow. (i) Participants face a wall during obstacle placement to avoid prior exposure. (ii) Upon the start signal, they observe the obstacles and draw HRZ using either a mouse (2D system) or hand gestures (MR system). (iii) After drawing, they press the ``Task Complete'' button, prompting the robot to navigate to the goal. (iv) The robot moves to the goal position.}
    \label{fig:experiment_method}
\end{figure}
To evaluate the effectiveness of the proposed MR-based Hand-drawn Map Editing system (MRHaD) for mobile robot navigation, we conducted a comprehensive within-subject study. The evaluation focused on the following performance criteria:
\begin{enumerate}
    \item \textbf{Navigation Success Rate:} The percentage of trials in which the mobile robot successfully reached the goal.
    \item \textbf{Map Reflection Accuracy:} The degree to which the drawn HRZ correctly represent actual obstacles on the navigation map.
    \item \textbf{Map Editing Time:} The time required to update the map with HRZ modifications.
    \item \textbf{Global Path Efficiency:} The reduction in the robot’s planned path length resulting from accurate HRZ placement.
    \item \textbf{Cognitive Load:} The mental effort required to comprehend and correlate the real-world environment with its digital map representation.
\end{enumerate}

\subsection{Experimental Method}\label{sec:EXP_previous_method}
Two drawing interfaces were implemented for comparison.
The proposed MR system (MRHaD) employs a HoloLens~2 to allow participants to create HRZ via natural hand gestures.
In contrast, the conventional 2D system is implemented on a laptop, where HRZ are drawn using mouse-based cursor control and button clicks.
Both systems offer equivalent functionality for HRZ creation, deletion, and clearing; they differ solely in the modality of interaction.

\subsection{Experimental Setup}
Fig.~\ref{fig:experiment_environment} illustrates the experimental setup, which consists of two obstacle configurations (Stage1 and Stage2).
In Stage1, a single flat obstacle is deployed, whereas Stage2 features a more challenging scenario with multiple obstacles.
The experimental procedure was as follows. Participants were first briefed on the study objectives and received practice only on the system (either 2D or MR) assigned for that session to avoid cross-system bias.

Fig.~\ref{fig:experiment_method} illustrates the overview of the experimental task flow.
With obstacles in place and to prevent prior knowledge of their positions, participants faced a wall during obstacle placement.
Upon receiving a start signal from the experimenter, participants observed the obstacles and then proceeded to enclose them by drawing HRZ using the assigned interface.
After completing the drawing task, they pressed the ``Task Complete'' button, which triggered the mobile robot’s navigation toward a predefined goal.
A five-minute break was provided before repeating the procedure with the alternate system.
Finally, participants completed questionnaires addressing both system-specific usability and overall experience.

A total of 16 participants (8 males and 8 females, aged 19–26) were recruited.
Survey responses indicated that 81.3\% had minimal experience with VR/MR head-mounted displays and 87.5\% had limited familiarity with Air-Tap gestures.
To mitigate potential learning effects, participants were evenly divided into two groups, with the order of system usage counterbalanced across subjects.

\subsection{Experimental Results}\label{sec:EXP_result}
\subsubsection{Navigation Success Rate}
\begin{table}[t]
    \centering
    \caption{Robot Navigation Results}
    \label{tab:navigation_results}
    \begin{tabular}{lcccc}
        \toprule
        & \multicolumn{2}{c}{Stage1} & \multicolumn{2}{c}{Stage2} \\
        \cmidrule(lr){2-3} \cmidrule(lr){4-5} 
        & \: 2D & MR \: & \: 2D & MR \: \\
        \midrule \midrule
        Success rate[\%] & 56.3 & 87.5 & 6.25 & 68.8 \\ \hline
        \textbf{Success} & \textbf{9} & \textbf{14} & \textbf{1} & \textbf{11} \\
        \textbf{Failure} & \textbf{7} & \textbf{2} & \textbf{15} & \textbf{5} \\
        \bottomrule
    \end{tabular}
\end{table}
\begin{figure}[t]
    \centering
    \includegraphics[keepaspectratio, width=0.97\linewidth]{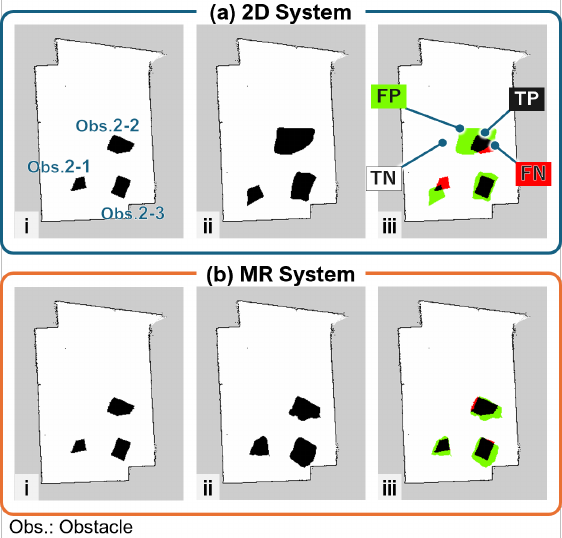}
    \caption{Examples of (i) the Ground Truth, (ii) the Drawn Map, and (iii) the Corresponding Pixel Classification. TP corresponds to black pixels except for wall areas, FP to green, FN to red, and TN to white.}
    \label{fig:map_comparison_example}
\end{figure}
Table~\ref{tab:navigation_results} shows the robot navigation results. In Stage1, the MR system achieved a success rate of 87.5\%, which was significantly higher than the 56.3\% success rate of the 2D system. In the more complex Stage2 configuration, the MR system attained a success rate of 68.75\%, while the 2D system’s performance dropped sharply to 6.25\%. These results indicate that HRZ created via the MR system more accurately reflects the real environment, thereby supporting more reliable path planning across varying levels of environmental complexity.

MR and 2D systems were analyzed using the Shapiro-Wilk test \cite{shapiro1965analysis} for normality.
Since normality was not observed in any of the evaluation metrics, the Wilcoxon signed-rank test \cite{rey2011wilcoxon} was used to examine statistical significance.
The experimental results are analyzed for each evaluation metric below.

\subsubsection{Map Reflection Accuracy}
Map reflection accuracy was quantitatively assessed by comparing the SLAM-generated map (Ground Truth) with the participant-generated HRZ map (Drawn Map). Each pixel was classified into four categories as illustrated in Fig.~\ref{fig:map_comparison_example}:
\begin{itemize}
    \item \textbf{True Positive (TP):} Cells classified as occupied in both the Ground Truth and the Drawn Map (excluding walls).
    \item \textbf{False Positive (FP):} Cells classified as free or unknown in the Ground Truth but classified as occupied in the Drawn Map.
    \item \textbf{False Negative (FN):} Cells classified as occupied in the Ground Truth but classified as free in the Drawn Map.
    \item \textbf{True Negative (TN):} Cells classified as free in both the Ground Truth and the Drawn Map.
\end{itemize}

\begin{table}[t]
    \centering
    \caption{Map Reflection Accuracy}
    \label{tab:map_difference}
    \scalebox{0.65}{
    \begin{tabular}{lccccccccccc}
        \toprule
        \multirow{2}{*}{\begin{tabular}{c}Stage/\\Obs.\textsuperscript{\dag}\end{tabular}} & & \multicolumn{2}{c}{Accuracy} & \multicolumn{2}{c}{Precision} & \multicolumn{2}{c}{Recall} & \multicolumn{2}{c}{Specificity} & \multicolumn{2}{c}{F-1 Score} \\
        \cmidrule(lr){3-4} \cmidrule(lr){5-6} \cmidrule(lr){7-8} \cmidrule(lr){9-10} \cmidrule(lr){11-12}
        &  & \: 2D & MR \: & \: 2D & MR \: & \: 2D & MR \: & \: 2D & MR \: & \: 2D & MR \: \\
        \midrule\midrule
        \multirow{2}{*}{Stage1} & $M$[\%] & 96.1 & 99.0 & 33.5 & 69.3 & 89.3 & 97.8 & 96.3 & 99.1 & 48.4 & 80.7 \\
        & $p$ & \multicolumn{2}{c}{$\boldsymbol{3.05\times10^{-5}}$} & \multicolumn{2}{c}{$\boldsymbol{3.05\times10^{-5}}$} & \multicolumn{2}{c}{\textbf{0.0182}} & \multicolumn{2}{c}{$\boldsymbol{6.10\times10^{-5}}$} & \multicolumn{2}{c}{$\boldsymbol{3.05\times10^{-5}}$} \\
        \midrule
        \multirow{2}{*}{Stage2} & $M$[\%] & 93.3 & 96.6 & 34.8 & 54.1 & 68.4 & 82.7 & 93.8 & 97.1 & 48.4 & 66.1 \\
        & $p$ & \multicolumn{2}{c}{$\boldsymbol{3.05\times10^{-5}}$} & \multicolumn{2}{c}{$\boldsymbol{6.10\times10^{-5}}$} & \multicolumn{2}{c}{0.0934} & \multicolumn{2}{c}{$\boldsymbol{3.05\times10^{-5}}$} & \multicolumn{2}{c}{$\boldsymbol{4.27\times10^{-4}}$} \\
        \midrule
        \multirow{2}{*}{Obs.2-1} & $M$[\%] & 98.5 & 99.5 & 26.5 & 61.9 & 73.7 & 94.1 & 98.8 & 99.6 & 39.7 & 72.8 \\
        & $p$ & \multicolumn{2}{c}{$\boldsymbol{3.05\times10^{-5}}$} & \multicolumn{2}{c}{$\boldsymbol{3.05\times10^{-5}}$} & \multicolumn{2}{c}{\textbf{0.0174}} & \multicolumn{2}{c}{$\boldsymbol{3.05\times10^{-5}}$} & \multicolumn{2}{c}{$\boldsymbol{3.05\times10^{-5}}$} \\
        \midrule
        \multirow{2}{*}{Obs.2-2} & $M$[\%] & 96.3 & 98.6 & 26.3 & 58.1 & 55.6 & 82.5 & 96.9 & 99.0 & 36.8 & 66.1 \\
        & $p$ & \multicolumn{2}{c}{$\boldsymbol{6.10\times10^{-5}}$} & \multicolumn{2}{c}{$\boldsymbol{6.10\times10^{-5}}$} & \multicolumn{2}{c}{\textbf{0.0214}} & \multicolumn{2}{c}{$\boldsymbol{6.10\times10^{-5}}$} & \multicolumn{2}{c}{$\boldsymbol{7.63\times10^{-4}}$} \\
        \midrule
        \multirow{2}{*}{Obs.2-3} & $M$[\%] & 98.3 & 98.5 & 44.8 & 47.2 & 93.5 & 79.3 & 98.3 & 98.7 & 59.5 & 59.8 \\
        & $p$ & \multicolumn{2}{c}{0.433} & \multicolumn{2}{c}{0.375} & \multicolumn{2}{c}{0.0507} & \multicolumn{2}{c}{0.117} & \multicolumn{2}{c}{0.821} \\
        \bottomrule
        \multicolumn{12}{l}{\textsuperscript{\dag} Obs.: Obstacle, $M$: Median}
    \end{tabular}
    }
\end{table}

Based on these classifications, the following metrics were calculated: Accuracy, Precision, Recall, Specificity, and F1-Score. Each metric represents the following:

\begin{itemize}
    \item Accuracy: The proportion of correctly classified regions in the Drawn Map compared to the Ground Truth. 
    \item Precision: The proportion of the restricted zones in the Drawn Map that correctly correspond to restricted zones in the Ground Truth.
    \item Recall: The proportion of restricted zones in the Ground Truth that were correctly reflected in the Drawn Map.
    \item Specificity: The proportion of navigable areas in the Ground Truth that were correctly maintained as navigable areas in the Drawn Map.
    \item F1-Score: The harmonic mean of Precision and Recall.
\end{itemize}

\begin{figure}[t]
    \begin{minipage}[b]{0.49\linewidth}
         \centering
         \includegraphics[keepaspectratio, width=1\linewidth]{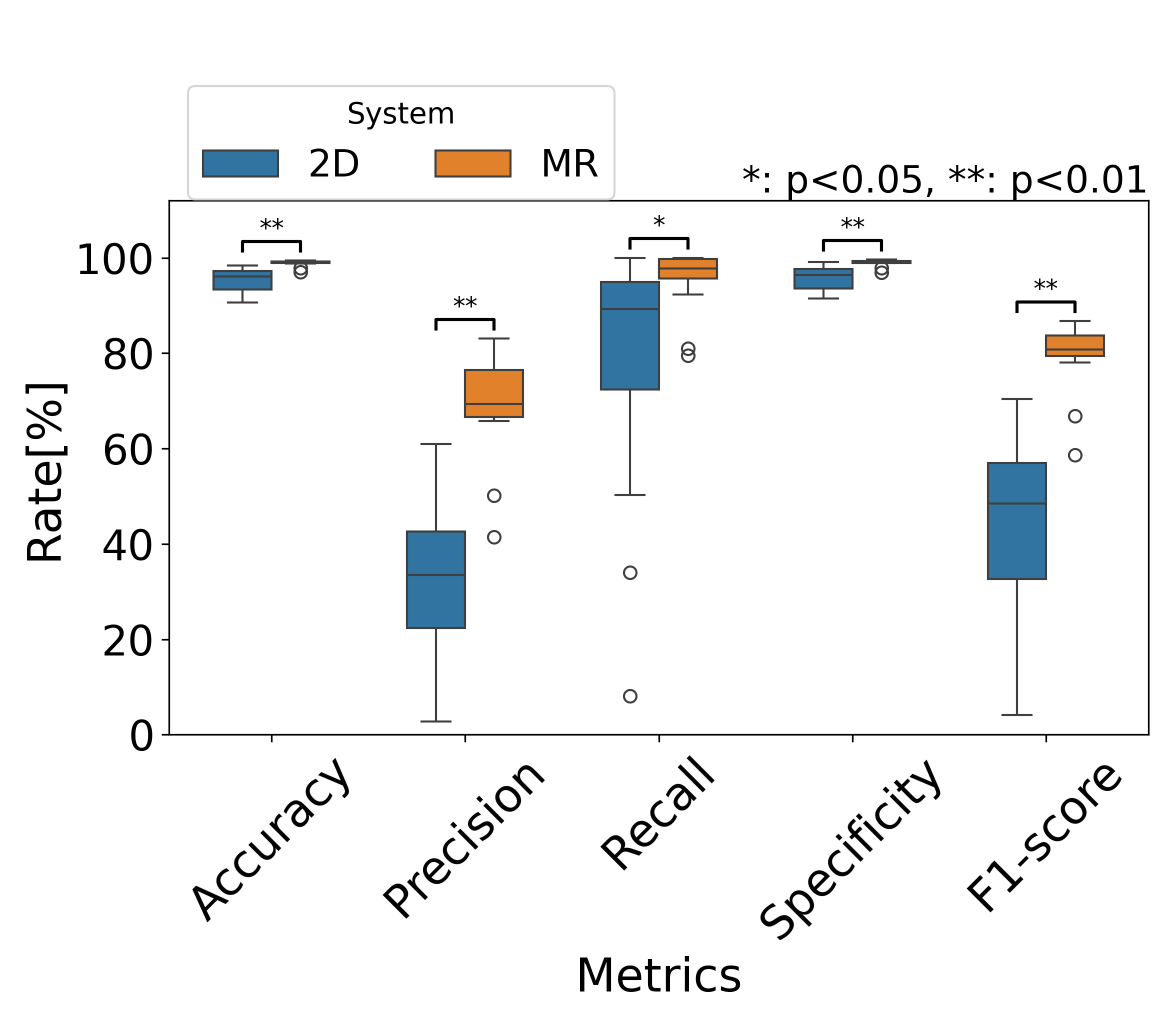}
         \subcaption{Stage1}
         \label{fig:map_difference_index_1}
    \end{minipage}
    \begin{minipage}[b]{0.49\linewidth}
         \centering
         \includegraphics[keepaspectratio, width=1\linewidth]{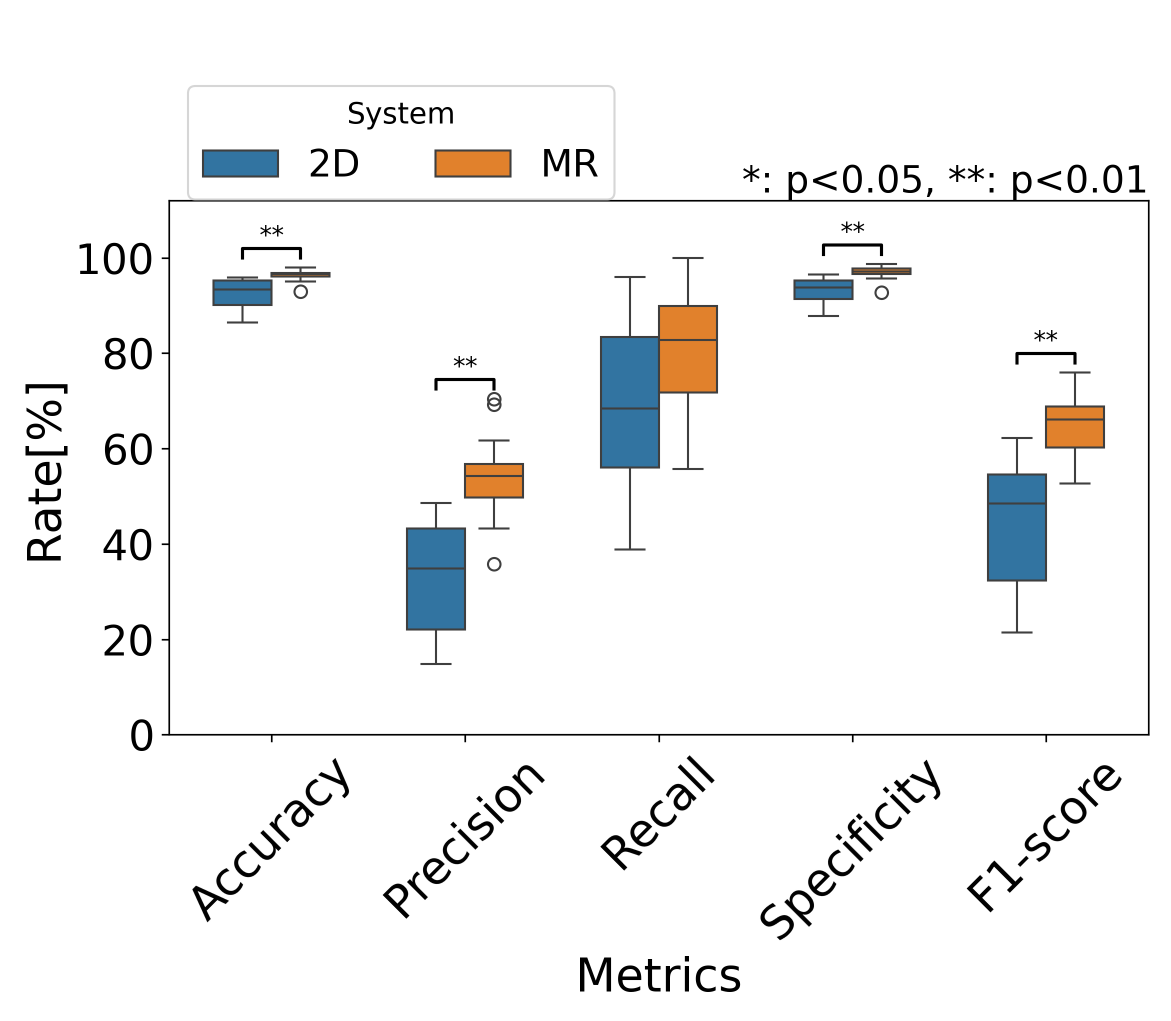}
         \subcaption{Stage2}
         \label{fig:map_difference_index_2}
    \end{minipage}
    \begin{minipage}[b]{0.49\linewidth}
         \centering
         \includegraphics[keepaspectratio, width=1\linewidth]{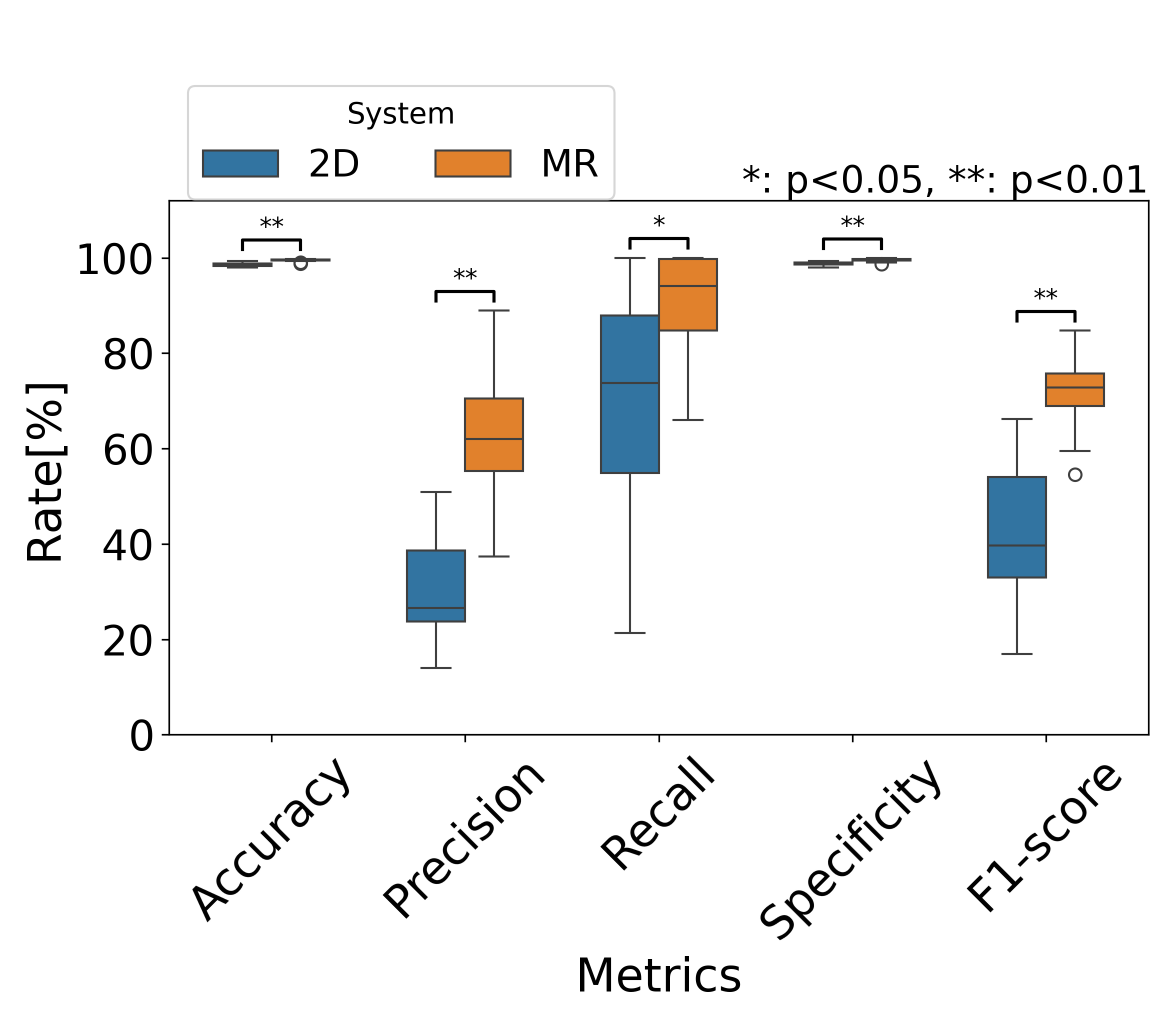}
         \subcaption{Stage2 - Only Obstacle2-1}
         \label{fig:map_difference_index_2-1}
    \end{minipage}
    \begin{minipage}[b]{0.49\linewidth}
         \centering
         \includegraphics[keepaspectratio, width=1\linewidth]{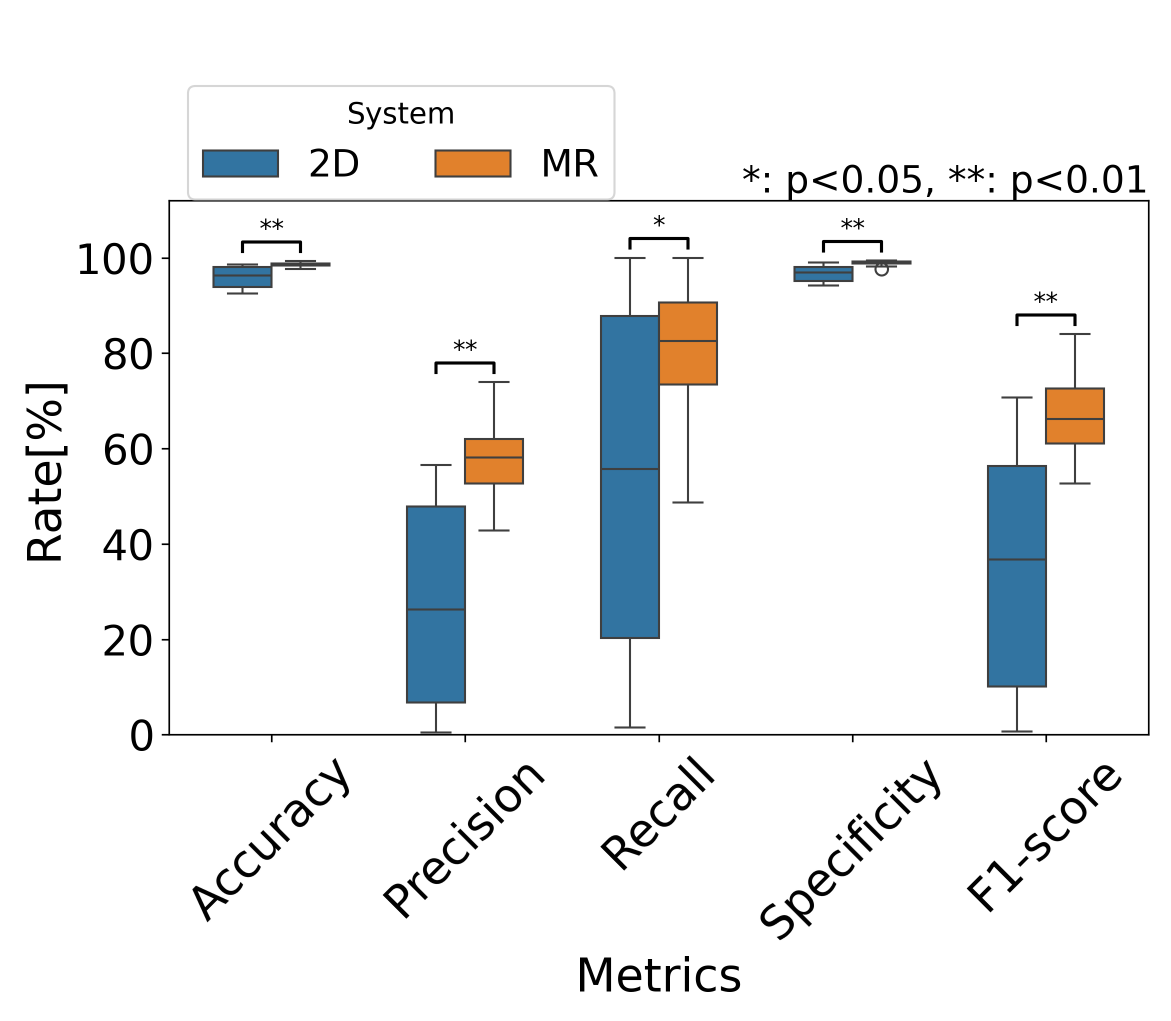}
         \subcaption{Stage2 - Only Obstacle2-2}
         \label{fig:map_difference_index_2-2}
    \end{minipage}
    \begin{minipage}[b]{0.49\linewidth}
         \centering
         \includegraphics[keepaspectratio, width=1\linewidth]{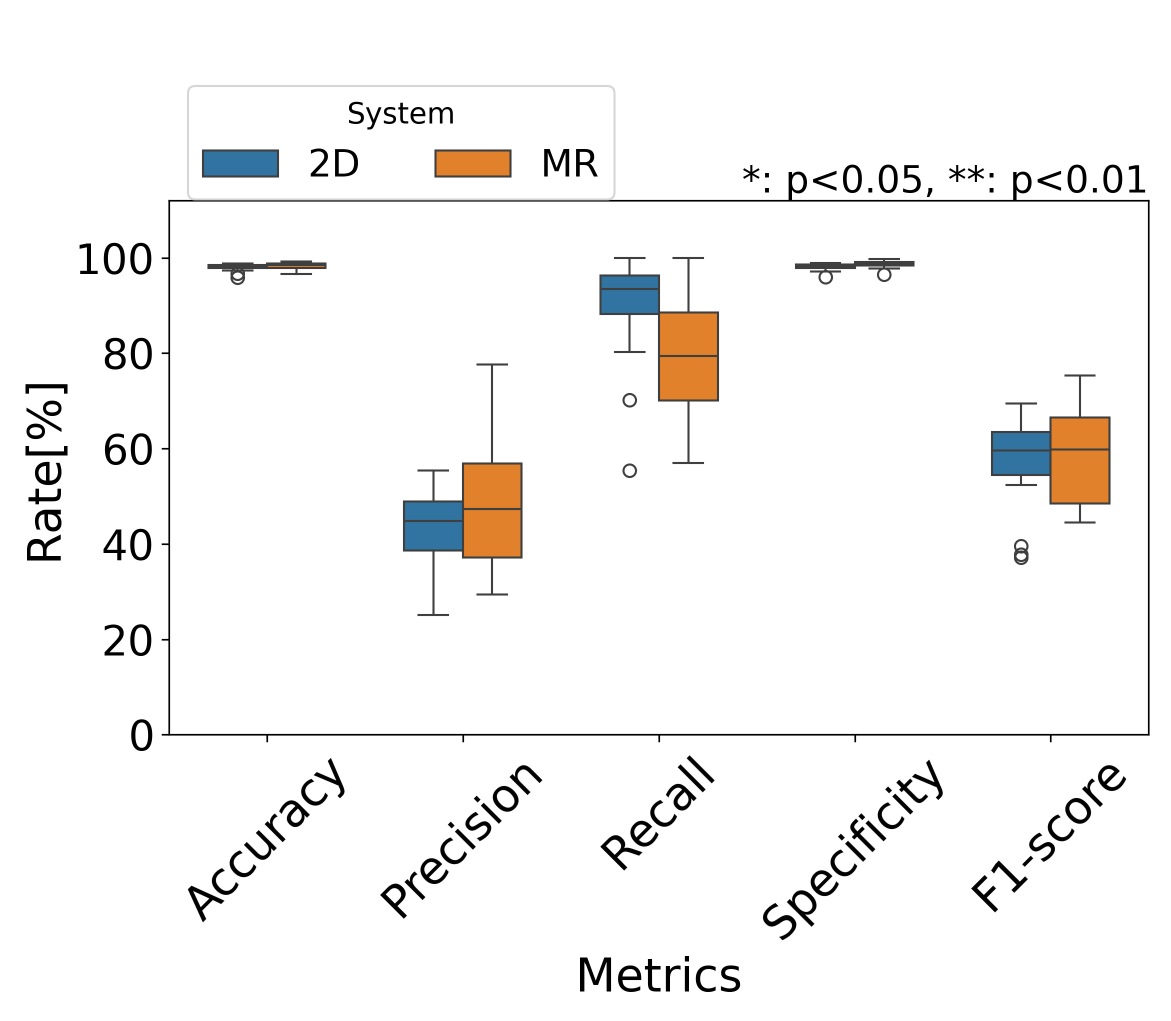}
         \subcaption{Stage2 - Only Obstacle2-3}
         \label{fig:map_difference_index_2-3}
    \end{minipage}
    \begin{minipage}[b]{0.49\linewidth}
         \centering
         \includegraphics[keepaspectratio, width=1\linewidth]{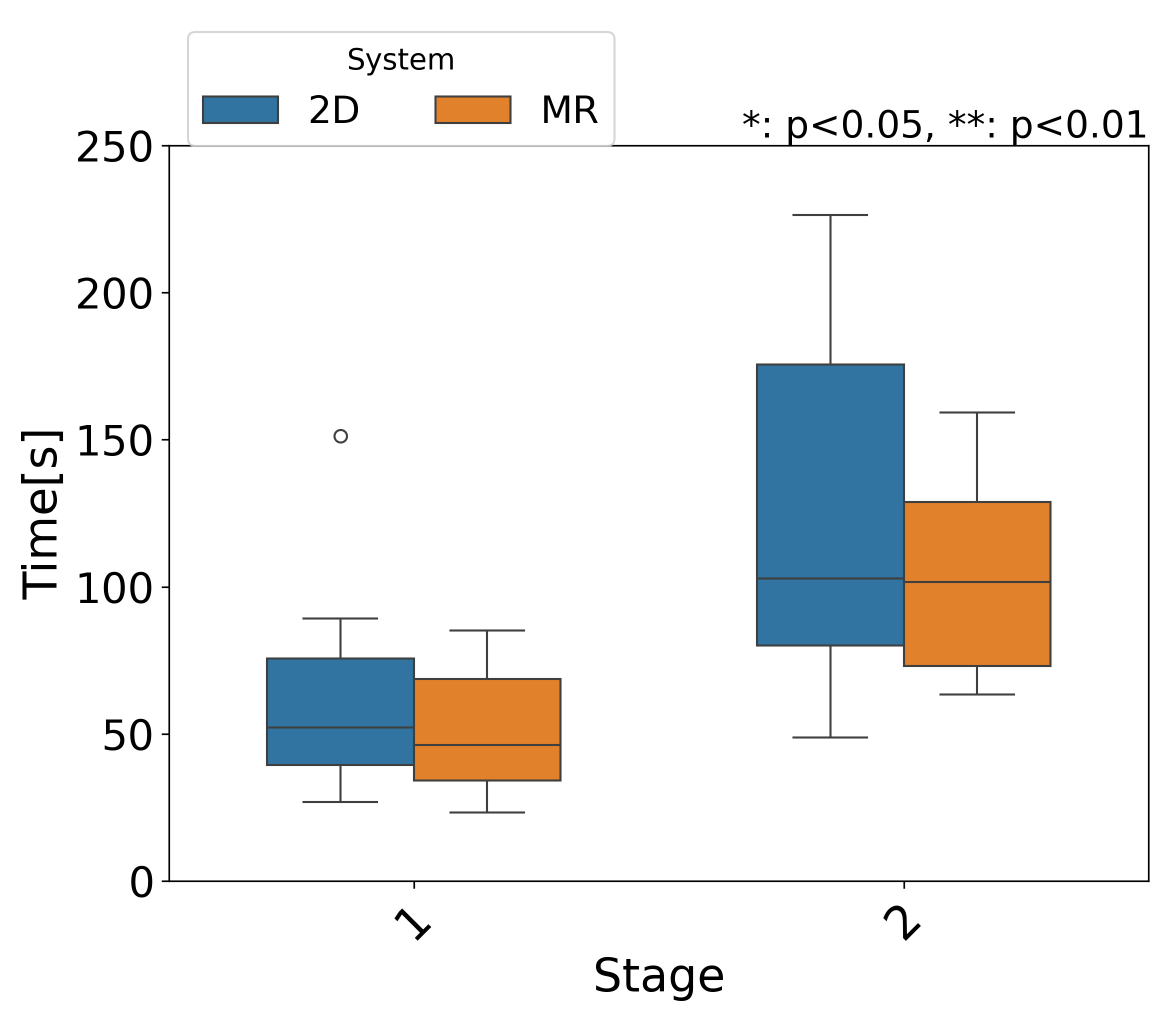}
         \subcaption{Task Completion Time}
         \label{fig:action_time_stage_boxplot}
    \end{minipage}
    \caption{Map Reflection Accuracy and Task Completion Time}
    \label{fig:map_difference_action_time}
\end{figure}
In autonomous mobile robot navigation, safety is the top priority, followed by efficiency. Therefore, Recall was given the highest priority in evaluation, followed by Precision and Specificity.
Table~\ref{tab:map_difference} and Fig.~\ref{fig:map_difference_action_time}(a)-(e) show the map reflection accuracy of the two systems.

In Stage1, Obstacle2-1, and Obstacle2-2, the MR system achieved superior scores across all metrics. 

In terms of Recall, the MR system outperformed the 2D system with 97.8\% compared to 89.3\% in Stage1, 94.1\% compared to 73.7\% in Obstacle2-1, and 82.5\% compared to 55.6\% in Obstacle2-2. These significant differences ($p<0.05$) suggest that the MR system has fewer omissions in reflecting the Ground Truth of the impassable area. On the other hand, in Obstacle2-3, the Recall of the 2D system was higher at 93.5\% compared to the MR system’s 79.3\%, which affected the overall Recall value for Stage2 and resulted in no statistically significant difference. The height of Obstacle2-3 and its distance from the participants likely made depth perception more challenging, resulting in positional errors during drawing with the MR system and insufficient coverage of the obstacle’s base.

Regarding Precision and Specificity, all obstacles except Obstacle2-3 showed higher values for the MR system compared to the 2D system, with statistically significant differences. These results indicate that the MR system has a lower rate of incorrectly marking areas that are actually passable as impassable, which means the MR system imposes fewer restrictions on the robot’s navigation range.

Furthermore, the MR system demonstrated higher median values with smaller variation among participants for Precision and Recall in Stage1 and Obstacle2-2, while the 2D system exhibited a broader range extending to lower values. A similar trend was observed in the F1-score, which is the harmonic mean of Precision and Recall.
This result is likely due to the fact that Obstacle1-1 and Obstacle2-2 had fewer corners or landmarks that could serve as positional references compared to other obstacles, resulting in larger drawing position errors.

These findings demonstrate that the MR system reduces the risk of the robot entering restricted zones while generating a map that more accurately represents the real environment.

\begin{figure}[t]
    \centering
    \begin{minipage}[b]{0.49\linewidth}
         \centering
         \includegraphics[keepaspectratio, width=1\linewidth]{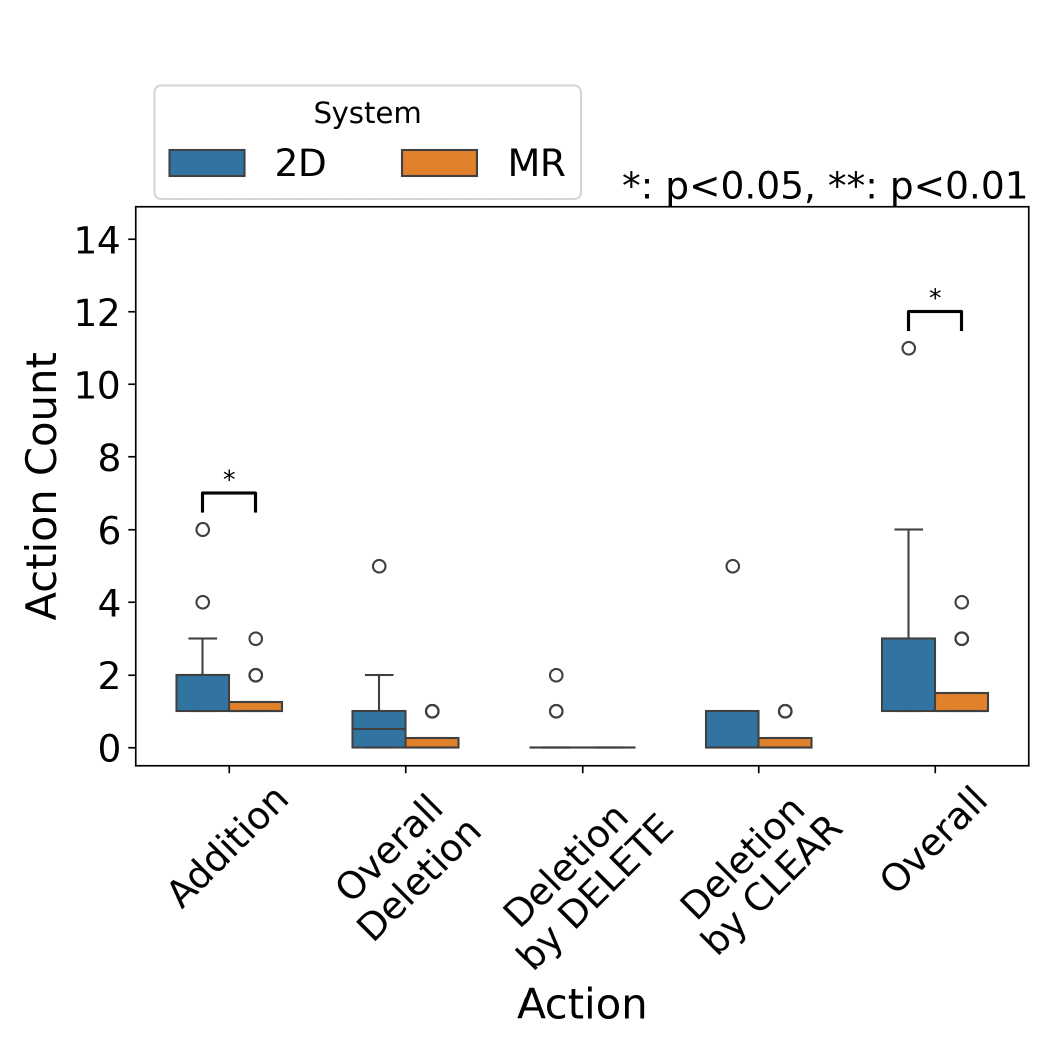}
         \subcaption{Stage1}
         \label{fig:drawing_number_boxplot_1}
    \end{minipage}
    \begin{minipage}[b]{0.49\linewidth}
         \centering
         \includegraphics[keepaspectratio, width=1\linewidth]{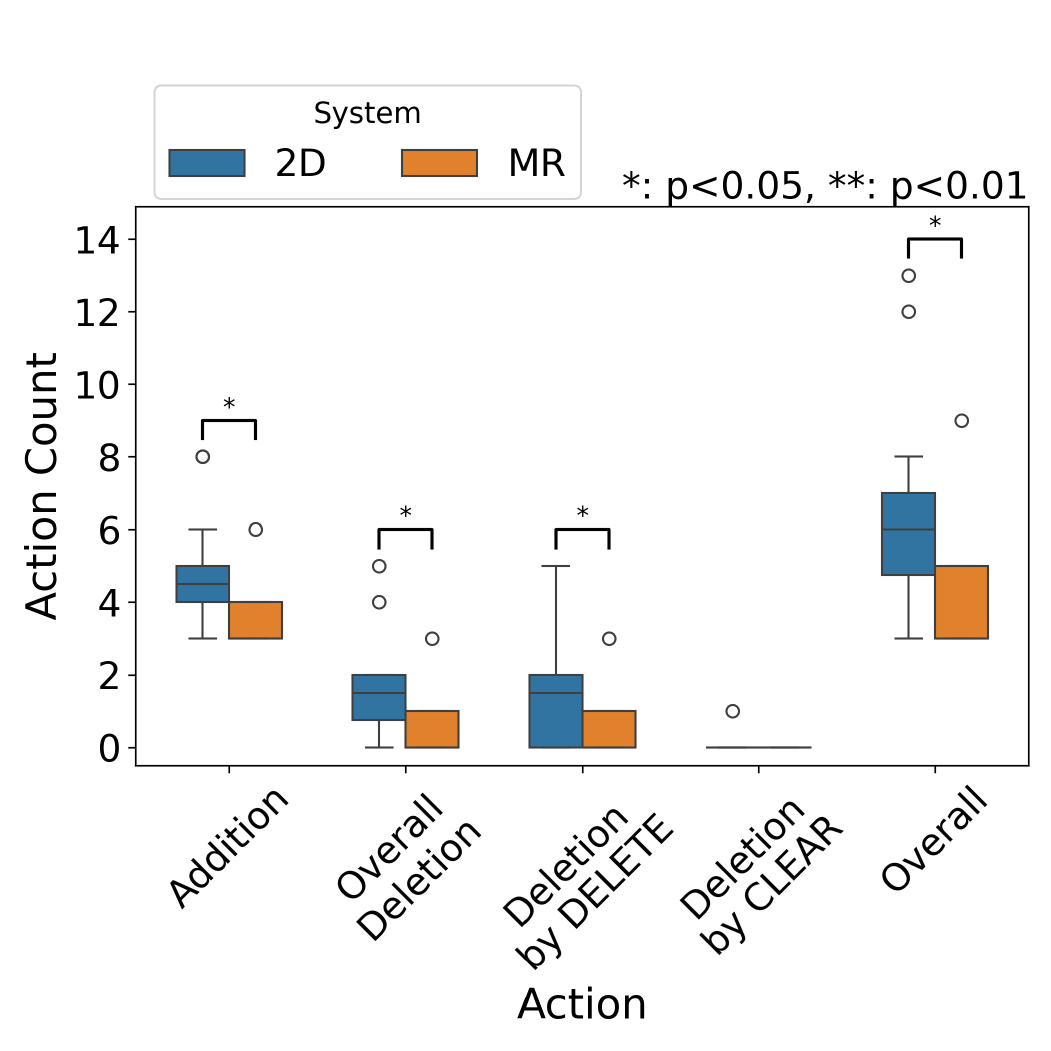}
         \subcaption{Stage2}
         \label{fig:drawing_number_boxplot_2}
    \end{minipage}
    \caption{The Number of Actions Required for Adding and Deleting HRZ.}
    \label{fig:drawing_number_boxplot}
\end{figure}
\begin{figure}[t]
    \centering
    \begin{minipage}{0.24\linewidth}
         \centering
         \includegraphics[keepaspectratio, width=\linewidth]{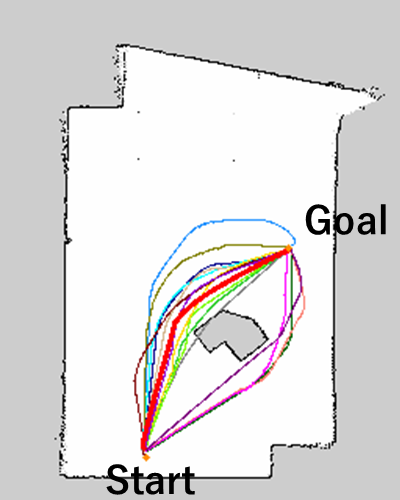}
         \subcaption{Stage1(2D)}
    \end{minipage}
    \begin{minipage}{0.24\linewidth}
         \centering
         \includegraphics[keepaspectratio, width=\linewidth]{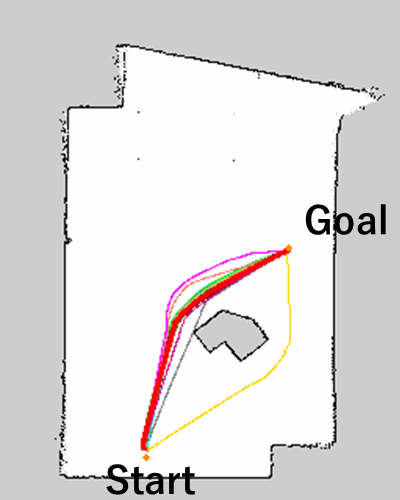}
         \subcaption{Stage1(MR)}
    \end{minipage}
    \begin{minipage}{0.24\linewidth}
         \centering
         \includegraphics[keepaspectratio, width=\linewidth]{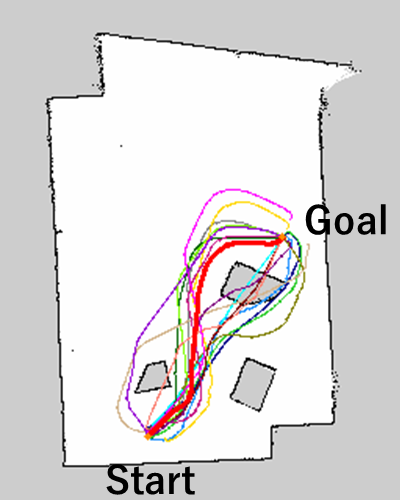}
         \subcaption{Stage2(2D)}
    \end{minipage}
    \begin{minipage}{0.24\linewidth}
         \centering
         \includegraphics[keepaspectratio, width=\linewidth]{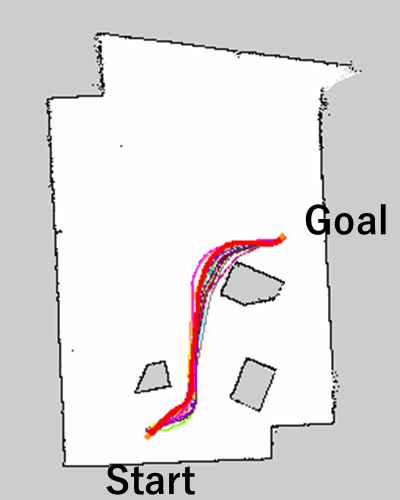}
         \subcaption{Stage2(MR)}
    \end{minipage}
    \caption{Results of Global Path Planning}
    \label{fig:global_path}
\end{figure}

\begin{table}[t]
    \centering
    \caption{Path Length of Global Path Planning}
    \label{tab:global_planning_path_length}
    \begin{tabular}{lllll}
        \toprule
        & \multicolumn{2}{c}{Stage1} & \multicolumn{2}{c}{Stage2} \\
        \cmidrule(lr){2-3} \cmidrule(lr){4-5} 
        & \: 2D & MR \: & \: 2D & MR \: \\
        \midrule
        $Median$ [\SI{}{m}] & 4.40 & 4.33 & 4.85 & 4.52 \\
        $Mean$ [\SI{}{m}] & 4.48 & 4.34 & 4.99 & 4.50 \\
        $SD$ [\SI{}{m}] & 0.400 & 0.106 & 0.949 & 0.127 \\
        \bottomrule
    \end{tabular}
\end{table}

\begin{figure}[t]
    \centering
    \begin{minipage}{0.44\linewidth}
         \centering
         \includegraphics[keepaspectratio, width=\linewidth]{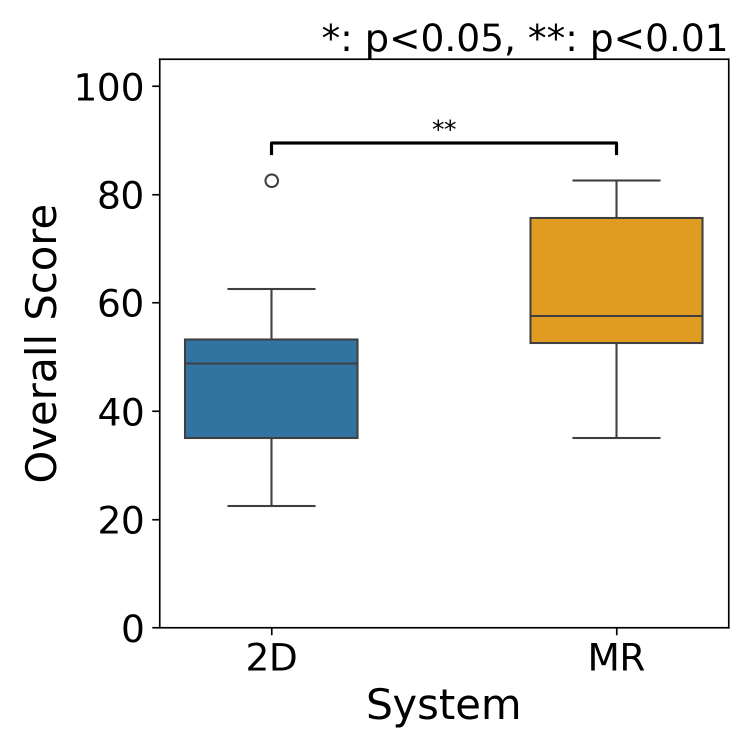}
         \subcaption{SUS}
         \label{fig:sus_overall_boxplot}
    \end{minipage}
    \begin{minipage}{0.44\linewidth}
         \centering
         \includegraphics[keepaspectratio, width=\linewidth]{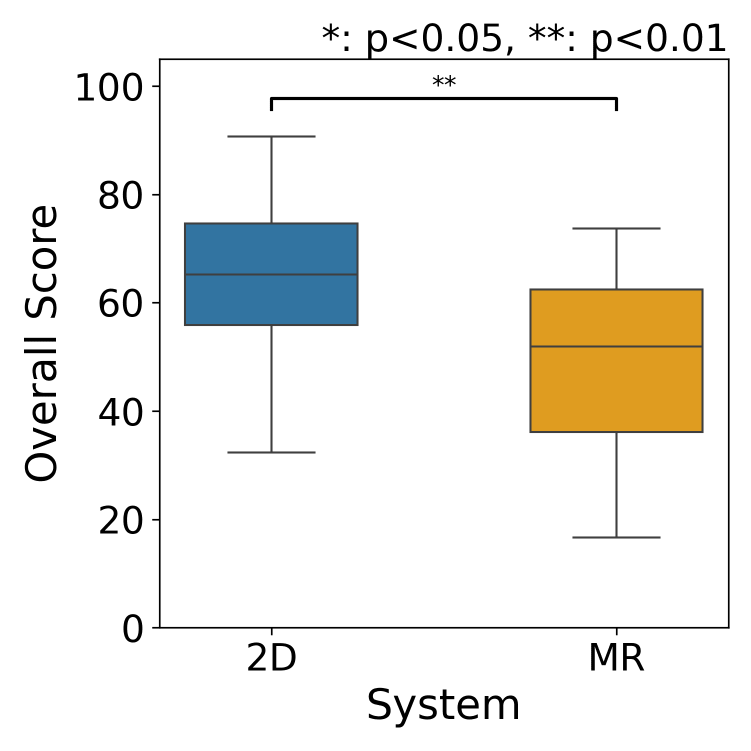}
         \subcaption{NASA-TLX}
         \label{fig:sus_metric_boxplot}
    \end{minipage}
    \caption{Score of SUS and NASA-TLX}
    \label{fig:sus_boxplot}
\end{figure}
\subsubsection{Map Editing Time}

Map editing speed was evaluated using task completion time, the number of actions for HRZ addition/removal, and function execution counts (ADD, DELETE, CLEAR).

Fig.~\ref{fig:map_difference_action_time}(f) shows the task completion times between the two systems. 
Although no statistically significant difference was observed in task completion time, the 2D system showed greater variation in task completion time among participants compared to the MR system.
Fig.~\ref{fig:drawing_number_boxplot} illustrates the number of drawing attempts required for each system. The MR system minimized drawing attempts, often allowing users to complete HRZ in a single attempt.
These results suggest that the MR system's visual feedback significantly improved operational speed.

\subsubsection{Global Path Efficiency}
The influence of HRZ accuracy on navigation was further examined via global path planning. Fig.~\ref{fig:global_path} displays the planned paths for both systems in Stage1 and Stage2.
Table~\ref{tab:global_planning_path_length} summarizes the path lengths, where the MR system produced marginally shorter median and mean path lengths with reduced variability.

These results indicate that improved map editing accuracy via MRHaD leads to more efficient and consistent global path planning.

\subsubsection{Cognitive Load}
Usability and mental workload were assessed using the System Usability Scale (SUS) \cite{brooke1996sus} and the NASA-TLX \cite{hart1988development}, respectively. As shown in Fig.~\ref{fig:sus_boxplot}, participants reported significantly higher SUS scores and lower NASA-TLX scores for the MR system compared to the 2D system, indicating reduced cognitive load and improved user satisfaction. Despite these benefits, several participants noted physical discomfort related to prolonged HoloLens~2 use and the execution of gesture-based interactions.
We posit that this was because most participants had limited experience with Air-Tap gestures, which may have contributed to the physical discomfort reported during prolonged HoloLens~2 use and gesture-based interactions.

\section{CONCLUSIONS}\label{sec:CON}
This paper presented MRHaD, MR-based intuitive map editing system that leverages hand-gesture interactions via HoloLens~2 to enable real-time modifications to navigation maps for mobile robots.
By directly linking the 2D map with the physical environment, MRHaD enhances the efficiency and safety of autonomous navigation compared to conventional 2D methods. Our experimental evaluation confirmed improvements in map reflection accuracy, reduced editing attempts, and lower cognitive load, thereby supporting more reliable robot navigation.
Overall, this study establishes the MR-based interface in robotics and human-robot interaction.

Nonetheless, our Stage2 results also reveal that the performance in complex scenarios is not yet optimal, indicating that further improvements are needed.
Future work should focus on enhancing gesture recognition robustness, refining the user interface to mitigate physical strain, and exploring supplementary control modalities to further improve navigation success in challenging environments.

% \section*{ACKNOWLEDGMENT}

\bibliographystyle{IEEEtran}

\end{document}